\documentclass[letterpaper, 10 pt, conference]{ieeeconf}  % Comment this line out if you need a4paper

\IEEEoverridecommandlockouts                              % This command is only needed if 
\usepackage{amssymb}  % assumes amsmath package installed

\UseRawInputEncoding
\usepackage[ruled,vlined,linesnumbered]{algorithm2e}
\usepackage{amsmath,amsfonts}
\usepackage{caption}
\usepackage{booktabs}
\usepackage{multirow}
\usepackage{makecell}
\usepackage{diagbox}
\usepackage{array}
\usepackage{booktabs}
\usepackage{bigstrut}
\usepackage{color}
\usepackage{xcolor}
\usepackage{colortbl}
\usepackage[colorlinks,
            linkcolor=green,
            anchorcolor=green,
            citecolor=green]{hyperref}
            
\usepackage{authblk}
\usepackage{array}
\usepackage[caption=false,font=normalsize,labelfont=sf,textfont=sf]{subfig}
\usepackage{textcomp}
\usepackage{stfloats}
\usepackage{url}
\usepackage{verbatim}
\usepackage{graphicx}
\usepackage{cite}
\usepackage{bbm}
\usepackage{hyperref}
\definecolor{mygreen}{rgb}{.75,1,.75}
\usepackage{etoolbox}
\makeatletter
\patchcmd{\@makecaption}
{\scshape}
{}
{}
{}
\newcommand{\customref}[2]{\hyperref[#2]{#1}}
\usepackage{pifont}

\usepackage[capitalize]{cleveref}
\makeatother

\title{\LARGE \bf
Fusion-Poly: A Polyhedral Framework Based on Spatial-Temporal Fusion for 3D Multi-Object Tracking  
}

\author{Xian Wu$^{\dagger}$, Yitao Wu$^{\dagger}$, Xiaoyu Li$^{\dagger}$, Zijia Li, Lijun Zhao$^{*}$, Lining Sun
	\thanks{$\dagger$ Equal contribution.}
    \thanks{$^*$ This work is supported by the Self-Planned Task (SKLRS202501E） of SKLRS(HIT) of China. {\textit{Corresponding author: Lijun Zhao.}}
    }
	\thanks{All authors are with the State Key Laboratory of Robotics and System, Harbin Institute of Technology, Harbin 150006, China. (E-mail: {\tt\small zhaolj@hit.edu.cn})}
}
\begin{document}

\maketitle
\thispagestyle{empty}
\pagestyle{empty}

%%%%%%%%%%%%%%%%%%%%%%%%%%%%%%%%%%%%%%%%%%%%%%%%%%%%%%%%%%%%%%%%%%%%%%%%%
\begin{abstract}
LiDAR-Camera 3D Multi-Object Tracking (MOT) approaches leverage rich visual features alongside precise depth, maintaining consistent motion states of surrounding obstacles and enhancing tracking reliability.
In practice, LiDAR and cameras operate at inherently different sampling frequencies. 
To ensure temporal alignment across these heterogeneous modalities, existing data pipelines typically synchronize multi-sensor streams and annotate them at a reduced shared frequency. 
Consequently, most prior approaches perform spatial fusion only at synchronized timestamps via projection-based or learnable cross-sensor association.
We argue that incorporating such asynchronous cues enables more frequent association and fusion, leading to more robust trajectory estimation over shorter temporal intervals.
To this end, we propose Fusion-Poly, a spatial-temporal fusion framework for 3D MOT that integrates asynchronous LiDAR and camera data. 
It allows trajectories to associate with multi-modal observations at synchronized timestamps (sync.) and with single-modal observations at asynchronous (async.) timestamps, enabling higher-frequency updates of motion and existence states.
Fusion-Poly includes a frequency-aware cascade matching module that adaptively handles sync. and async. frames according to upstream detection modalities, and a frequency-aware trajectory estimation module that maintains trajectory states through high-frequency motion prediction, differential updates, and confidence-calibrated lifecycle management. 
In addition, we design a full-state observation alignment module based on image-projection error optimization to enhance cross-modal consistency at sync. timestamps.
Fusion-Poly achieves \textcolor{black}{76.5\%} AMOTA on the nuScenes test set, the state-of-the-art performance among 3D MOT methods based on the Tracking-By-Detection (TBD) paradigm. 
Ablation studies further validate the effectiveness of each component in the proposed framework.
The code is available at \href{https://github.com/K544-AD/Fusion-Poly}{GitHub}.

% 是否 visual 待定
% To enable tracks to associate with both multi-modal and single-modal data at sync. and async. timestamps, respectively, and update the motion states and existence states of trajectories at a higher frequency, we propose two key components: a multi-stage association strategy adaptable to both data by flexibly switching association stages based on synchronization status, and \textcolor{black}{a confidence-based lifecycle management module that updates trajectories states by fully leveraging the predictive insights from 2D and 3D detectors at both timestamps.}
% \textcolor{black}{We propose two key components to achieve this: a context-aware adaptive multi-phase association paradigm that dynamically adapts to synchronization status to accommodate both data types, and a modality-informed confidence-calibrated lifecycle management module that updates trajectory states by leveraging dual-sensor predictive insights across both timestamp types.}
% 对于前者，a frequency-aware cascade matching module that adaptively fits synchronous/async. frames based on upstream detection modalities
% 对于后者，a frequency-aware trajectory estimation module that maintains trajectory states via high-frequency motion update, differential updates existence states of trajectory via confidence-calibrated lifecycle management (motion, existence states)
% On the KITTI dataset, Fusion-Poly exhibits competitive accuracy with \textcolor{black}{XX\%} HOTA.

\end{abstract}

% \begin{abstract}
% LiDAR-Camera 3D Multi-Object Tracking (MOT) leverages complementary sensor modalities to maintain consistent object states. 
% However, existing methods typically perform spatial fusion only at sync. timestamps, largely neglecting the high-frequency async. data inherent to divergent sensor sampling rates. 
% We argue that incorporating such async. cues can further improve robust trajectory estimation over short temporal horizons.
% To this end, we propose Fusion-Poly, a unified framework that seamlessly integrates synchronous and async. observations. 
% First, to ensure precise spatial consistency, we introduce a Geometry-Aware Alignment Module (GAAM) that refines 3D detections via full-state optimization based on geometric projection errors. 
% Second, to achieve robust spatiotemporal fusion, we design a frequency-aware mechanism comprising two synergistic modules: a Frequency-Aware Cascade Matching (FACM) module and a Frequency-Aware Trajectory Estimation (FATE) module. 
% This mechanism adaptively coordinates multi-stage association across synchronous and async. frames while maintaining high-frequency trajectory states through differential updates and confidence-calibrated lifecycle management. 
% Fusion-Poly achieves state-of-the-art performance among Tracking-By-Detection methods with XX.X\% AMOTA on the nuScenes benchmark. 
% Comprehensive ablation studies further validate the efficacy of our frequency-aware design. 
% The code will be released.
% \end{abstract}

\section{INTRODUCTION}
3D Multi-Object Tracking (MOT) is crucial for maintaining temporal consistency and estimating the motion trajectories of surrounding objects. 
Compared to single-modal approaches~\cite{li2023poly,li2024fast}, LiDAR-Camera MOT methods~\cite{caesar2020nuscenes,geiger2012_kitti,sun2020scalability_waymo} achieve superior performance by fusing dense semantic information with precise depth.
In practice, LiDAR and camera sensors operate at different sampling frequencies. 
To ensure temporal alignment across modalities, existing data infrastructures (Waymo~\cite{sun2020scalability_waymo}, nuScenes~\cite{caesar2020nuscenes}, etc.) typically perform multi-sensor synchronization and annotation at a lower unified frequency. 
Consequently, most prior methods~\cite{kim2021eagermot,xu2024emmsmot,wang2022deepfusionmot} restrict tracking to sync. timestamps, largely overlooking the high-frequency async. observations arising from heterogeneous sensor rates.
High-frequency sensor data has been shown to improve tracking performance~\cite{pang2022simpletrack}, as async. observation streams enable more frequent association and fusion, leading to more accurate tracklet estimation over short temporal intervals. 
Therefore, effectively integrating async. sensor data is critical for multi-modal 3D MOT.

% This is attributed to the fact that motion states are more predictable within short time intervals, mitigating the effects of occlusion through leveraging async. visual streams
% Furthermore, objects maintain persistent existence and do not exhibit abrupt motion variations when subjected to occlusion.
% Thus, high-frequency observations are conducive to mitigating the effects of occlusion through leveraging async. visual streams, which boosts observation rates.
% Specifically, async. visual data delivers abundant discriminative features to support accurate 2D detection, enabling the system to capitalize on high-frequency updates.
% This not only strengthens the algorithm’s occlusion resistance but also furnishes robust temporal features for target association.

\begin{figure}[t]
      \centering
      \includegraphics[width=\linewidth]{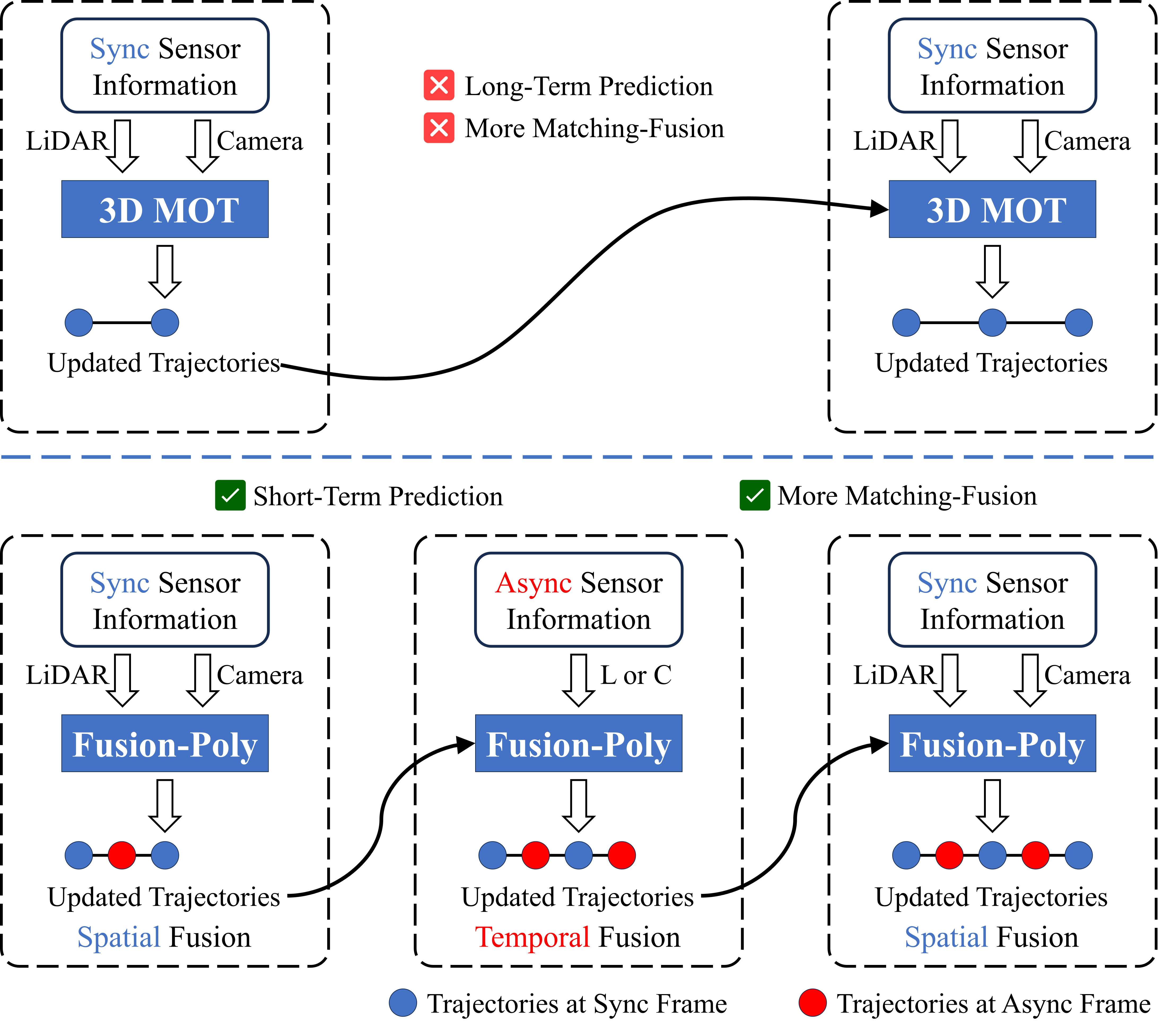}
      \vspace{-2em}
      \caption[l]{
      % Difference between LiDAR-Camera 3D MOT and Fusion-Poly. 
      % Compared with other methods, Fusion-Poly accounts for asynchrony arising from inconsistent sensor frequencies. 
      % By introducing asynchronous unimodal data, it increases the frequency of trajectory-observation matching and fusion, thus improving tracking robustness.
      Fusion-Poly differs from existing LiDAR-Camera 3D MOT methods by explicitly exploiting asynchronous sensor data at different frequencies, increasing matching frequency, and improving tracking robustness.
      }
    \label{fig:tease}
    \vspace{-1em}
\end{figure}

% how to perform temporal fusion, insight (What additional information do different frequencies bring?) -> How and why do we use (why camera-centric in the non-keyframe?, 除了camera本身的频率高之外，是不是camera好的泛化性也是我们这么做的理由) -> the performance and advantages (w/o additional training, etc.)
% In the data association module, we propose an adaptive three-stage association strategy to handle camera-only, LiDAR-only, and mixed-modality observation scenarios in a unified manner. 
% Specifically, when multi-modal observations are available from upstream, the module executes the full three-stage association procedure, whereas in single-modality cases, it automatically degenerates to a single-stage association. 
% We define keyframes as timestamps with sync. sensor data, and non-keyframes as those featuring async. data. 
% We then focus on trajectories to correlate and fuse data across these two types of frames, a process referred to as temporal fusion. 
% Similarly, we define spatial fusion as the interaction among trajectories, 2D detections, and 3D detections at the same timestamp, which aims to yield a unified instance representation.
% However, under key frames and non-key frames, the modalities of incoming detection results from upstream detection are inconsistent. 
% We design a unified spatial-temporal fusion strategy to enable trajectories to be associated and fused with observations from different sensors under both synchronous and async. timestamps through XXX association and XXX life management.

To this end, we propose Fusion-Poly, a polyhedral spatial-temporal fusion framework for 3D MOT that integrates async. LiDAR and camera data. To leverage multi-sensor observations under both sync. and async. timestamps, we design a unified spatial-temporal fusion mechanism composed of dedicated matching and trajectory estimation modules. The former enables high-frequency trajectory-observation association, while the latter maintains motion and existence states.
We introduce a Frequency-Aware Cascade Matching Module (FACM) that dynamically coordinates tracklets with heterogeneous observations. It adaptively performs modality-specific association strategies according to upstream detections, enabling flexible and efficient matching across sync. and async. frames while fully exploiting multi-modal cues.
We further propose a Frequency-Aware Trajectory Estimation Module (FATE). For trajectory existence modeling, we develop a modality-informed, confidence-calibrated lifecycle management strategy that evaluates trajectory quality based on multi-modal observation scores from both sync. and async. frames. This trajectory-centric spatial-temporal fusion mechanism integrates multi-modal information spatially and exploits async. observations temporally, improving motion accuracy and tracking reliability.

To enhance spatial consistency between LiDAR and camera detections, we additionally introduce a Geometry-Aware Alignment Module (GAAM), which projects 3D bounding boxes onto the image plane and performs full-state optimization using IoU-based residuals.

% This approach jointly refines the complete set of 3D state variables, including localization, size and orientation. 
% It thus yields an accurate and spatially consistent 3D observation.}

Fusion-Poly is learning-free and follows the TBD paradigm, allowing seamless integration with various detectors. 
With the Python implementation, Fusion-Poly is extensively evaluated on nuScenes, a large-scale autonomous driving benchmark, achieving \textcolor{black}{76.5\%} AMOTA on the nuScenes test set. The code will be open-source to contribute to the community.
Our contributions include four aspects:

\begin{itemize}

\item We propose Fusion-Poly, a unified LiDAR-Camera 3D MOT framework that jointly performs cross-modal fusion and cross-frequency integration.

\item We introduce GAAM, a geometry-aware alignment module that enhances spatial fusion through joint 2D-3D optimization.

\item We develop two frequency-aware components, FACM and FATE, which enable high-frequency trajectory association and state estimation under both sync. and async. settings.

\item Fusion-Poly achieves SOTA among TBD methods with 76.5\% AMOTA on nuScenes. 

\end{itemize}

\section{RELATED WORK}

\textbf{Single-Modal 3D MOT.}
% overall tracking pipeline
Traditional 3D MOT methods typically follow a Tracking-By-Detection framework~\cite{pang2022simpletrack, benbarka2021score, li2023poly, li2024fast}, leveraging off-the-shelf detectors as inputs, followed by motion estimation, data association, and a trajectory life management module. 
% 早期方法通常是lidar-only, xx好，cite some methods
% LiDAR-only methods have been widely adopted in 3D MOT due to their accurate geometric measurements and robustness to lighting conditions.
Early 3D detectors were dominated by LiDAR due to its ability to provide accurate geometric measurements; consequently, early TBD tracking research~\cite{li2023poly, li2024fast, pang2022simpletrack, weng_ab3dmot} predominantly focused on LiDAR-based solutions.
However, LiDAR-only methods struggle in highly dynamic or cluttered environments due to point cloud sparsity and the difficulty in distinguishing shape-similar objects lacking appearance cues.
With the development of deep learning, camera-based methods~\cite{zhang2022mutr3d,doll2023star,ding2024ada,kappeler2025bridging, ding2025tqd, yu2026g2dp,li2026HAT} have achieved significant progress due to their ability to capture rich texture information.
% 2-3 methods
These methods typically achieve cross-temporal association of targets through a cross-attention mechanism and realize end-to-end MOT and show great potential.
Despite these advancements, camera-only approaches remain limited by depth ambiguity, trailing LiDAR-based counterparts in tracking precision.

\textbf{LiDAR-Camera 3D MOT.}
By integrating precise depth from LiDAR with rich semantics from cameras, multi-modal tracking has become the dominant paradigm~\cite{kim2021eagermot, zeng2024fusiontrack}.
Existing approaches generally fall into two categories: result-level fusion and feature-level fusion.
Result-level methods, such as EagerMOT~\cite{kim2021eagermot} and DeepFusionMOT~\cite{wang2022deepfusionmot}, focus on associating 2D and 3D detections to enhance state estimation.
Notably, EMMS-MOT~\cite{xu2024emmsmot} further refines 3D positions by minimizing the projected Euclidean distance between cross-modal boxes.
Feature-level methods, such as CAMO-MOT~\cite{li2023camo} and DINO-MOT~\cite{lee2024dino}, leverage visual features or occlusion heads to improve association robustness.
However, these methods predominantly operate on sync. frames, assuming simultaneous availability of multi-sensor data.
This reliance neglects the high-frequency async. data inherent to heterogeneous sensors.
In contrast, Fusion-Poly introduces a unified spatial-temporal framework that effectively exploits both sync. and async. data, unlocking the potential of high-frequency tracking.

%%%%%%%%%%%%%%%%%%%%%%%%%%%%%%%%%%%%%%%%%%%%%%%%%%%%%%%%%%%%%%%%%%%%%%%%%%%%%%%%

\begin{figure*}
    \centering
    \includegraphics[width=1\linewidth]{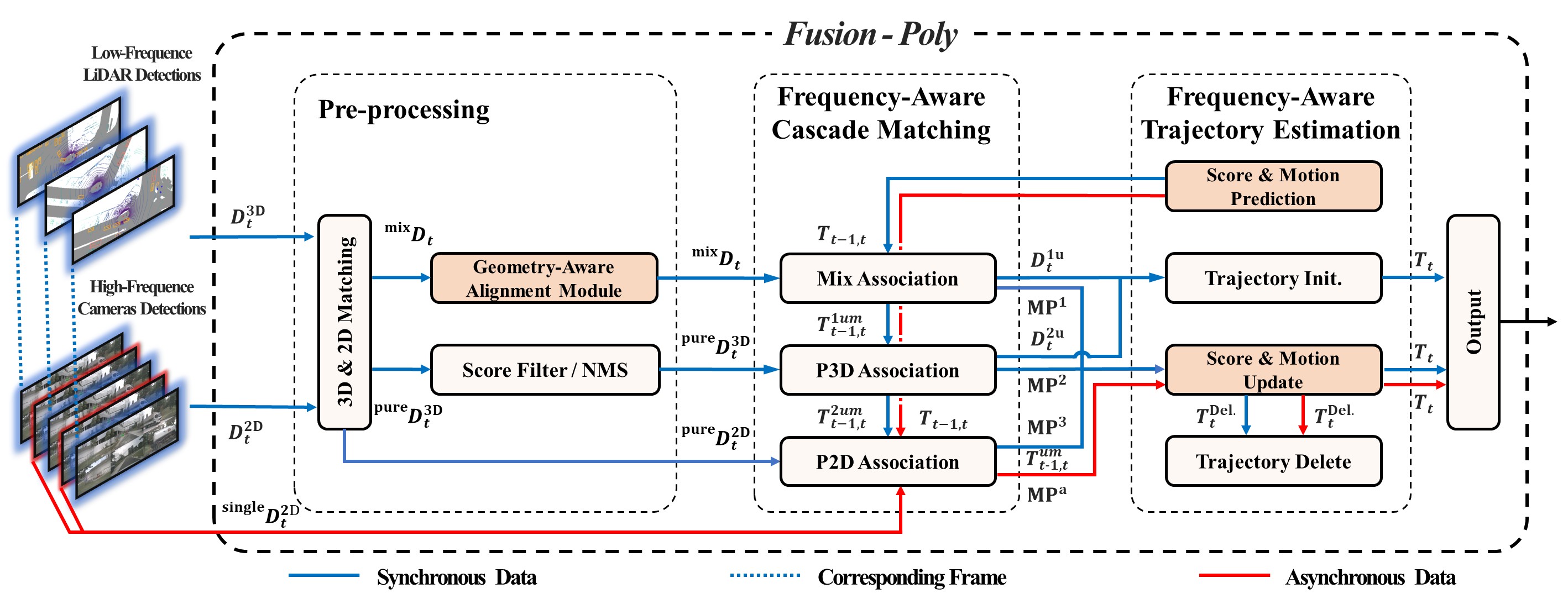}
    \caption{The pipeline of Fusion-Poly. We propose a unified LiDAR-Camera 3D MOT framework that jointly performs cross-modal fusion and cross-frequency integration by simultaneously incorporating synchronous and asynchronous data. % achieving comprehensive spatial-temporal fusion.
    }
    \label{fig:pipeline}
    \vspace{-1em}
\end{figure*}

\section{METHOD}
% spatial and temporal multi-modal
% In this section
% In this section, we present the overall architecture of Fusion-Poly and detail each of its components.
% We propose Fusion-Poly, a holistic spatial-temporal fusion 3D MOT framework that performs both multi-modality and varied-frequency fusion.
% The overall framework is illustrated in~\ref{fig: tease}, and detailed descriptions of each component are provided below.

% \textcolor{black}{In this section}, we present the overall pipeline of Fusion-Poly and describe each of its components in detail.

\subsection{\textcolor{black}{Overall Architecture}}
% \textcolor{black}{How pre-processing, matching, and trajectory estimation work in async/sync data flow.}
% input (dual stream), fusion-poly通过xx, xx, xx利用这些信息。
% 主旨：概括全文+引人注意，不要太技术，也要讲差异和见解，主次分明（写作的主语和宾语），跑出来后面编的module name
% 可以讲后面section的第一段精炼成1-2句话放在这块
% 首先/然后/最后，我们使用xx module来xx(impletation and output)，xx了(performance and insight)
% outp
% \textcolor{black}{Fusion-Poly takes LiDAR-Camera dual-stream asynchronous detections as input.}
Fusion-Poly takes async. LiDAR and camera detections with disparate sampling frequencies as inputs.
First, to enhance the spatial consistency of sync. detections, we propose a Geometry-Aware Alignment Module (GAAM, \cref{sec:gaam}), \textcolor{black}{which operates by minimizing re-projection errors after detection-level matching.}
% First, to improve the spatial consistency between LiDAR and camera modalities, we propose the Geometry-Aware Alignment Module (\cref{sec:gaam}) \textcolor{black}{for cross-modal spatial alignment} that refines 3D detections.
% minimizing the discrepancy between projected 3D bounding boxes and 2D detections
We then propose a Frequency-Aware Cascade Matching Module (FACM, \cref{sec:faca}) to dynamically coordinate the correlation between trajectories and heterogeneous observations \textcolor{black}{by flexibly switching matching strategies based on the synchronization status of frames}.
Finally, \textcolor{black}{to achieve more robust trajectory state estimation over shorter temporal intervals}, we propose a Frequency-Aware Trajectory Estimation Module (FATE, \cref{sec:fatep,sec:fateu}). 
This module performs differential motion estimation for trajectory motion states, and adopts a modality-informed, confidence-calibrated lifecycle management strategy for trajectory existence states.
It effectively mitigates the unreliability induced by async. observations while exploiting the useful informative cues.
Fusion-Poly ultimately produces smooth trajectories for 3D objects.

\textcolor{black}{\subsection{Pre-processing}}
\label{sec:gaam}
The pre-processing module improves the quality of 3D detections fed into the tracking pipeline.
It comprises two sub-modules: matching and alignment. 
Specifically, the matching process establishes correspondences between 3D LiDAR detections and 2D camera detections.
The alignment sub-module then refines the corresponding 3D detections by leveraging 2D camera detection cues.
The inputs are categorized into two groups: LiDAR 3D and camera 2D detections at low-frequency sync. timestamps, and single-modality camera 2D detections from high-frequency streams.
% Pre-processing overall pipeline, matching and alignment, matching是什么意思，怎么干的，为了什么。alignment是什么意思，怎么干的，为了什么

% matching-only pre-processing -> 不全面的空间一致性,2D 检测是很准的，
% Existing methods~\cite{wang2022deepfusionmot, li2023camo, kim2021eagermot} 利用 spatial consistency between LiDAR and camera modalities only employ a matching-only Pre-Processing strategy to obtain 3D-2D detection pairs.
% 这是一种不全面的空间一致性利用，
% To improve the spatial consistency between these two detection modalities.
% Yet this approach inevitably contaminates the tracking pipeline with low-quality 3D detections.
% To address this issue, we propose a \textbf{Geometry-Aware Alignment Module}, an IoU-based optimization framework for cross-modal spatial alignment that refines 3D detections.
% Specifically, this module first establishes 3D-2D detection correspondences, then optimizes detection quality by minimizing the discrepancy between the projected 3D bounding box and the 2D detection via an IoU-based objective.

% \textbf{Input.}

% The input stream type is dictated by whether the current frame constitutes a multi-modal synchronous frame.

\textbf{Matching.}
% \textcolor{black}{In each frame}, 
% To obtain the inter-sensor observation correspondences for the same object in a synchronous frame, we first perform matching on frames with multi-sensor \textcolor{black}{data flow}.
To establish inter-sensor observation correspondences for the same object within a sync. frame, we first match across multi-sensor inputs.
This process takes 3D detections \(D^\text{3D}_t = (x, y, z, w, l, h, \theta, s, \text{cls})\) and 2D detections \(D^\text{2D}_t = (x_1, y_1, x_2, y_2, s, \text{cls})\) from off-the-shelf detectors as input. 
\((x, y, z)\) denotes the 3D geometric center, \((w, l, h)\) represents the object dimensions (width, length, height), \(\theta\) is the heading angle, \(s\) stands for the detection confidence score, \(\text{cls}\) denotes the object category. 
For 2D detections, \((x_1, y_1)\) and \((x_2, y_2)\) correspond to the top-left and bottom-right corners of the bounding box, respectively.
% This process takes 3D detections $D^\text{3D}_t = \{d^\text{3D}_i\}{i=1}^{N^\text{3D}t}$ and 2D detections $D^\text{2D}_t = \{d^\text{2D}_j\}{j=1}^{N^\text{2D}t}$ from off-the-shelf detectors as input. Each 3D detection $d^\text{3D}_i = (x, y, z, w, l, h, \theta, s, \text{cls})$ consists of the 3D center coordinates $(x, y, z)$, object dimensions $(w, l, h)$, heading angle $\theta$, confidence score $s$, and category label $\text{cls}$. Each 2D detection $d^\text{2D}_j = (x_1, y_1, x_2, y_2, s, \text{cls})$ is represented by the top-left $(x_1, y_1)$ and bottom-right $(x_2, y_2)$ corners of the bounding box, along with the confidence score $s$ and category label $\text{cls}$.

We project the $D^\text{3D}_t$ onto the image plane and perform IoU-based matching with $D^\text{2D}_t$.
% This results yields three categories: 3D-2D mix detections ${}^{{\text{mix}}}D_{t}$, which consist of ${}^{{\text{mix}}}D^{\text{3D}}_{t}$ and ${}^{{\text{mix}}}D^{\text{2D}}_{t}$; pure 3D detections ${}^{{\text{pure}}}D^{\text{3D}}_{t}$; and pure 2D detections ${}^{{\text{pure}}}D^{\text{2D}}_{t}$.
% For ${}^{{\text{pure}}}D^{\text{3D}}_{t}$, we apply Score Filter (SF) and Non-Maximum Suppression (NMS) to discard unreliable results.
% \textcolor{red}{${}^{{\text{pure}}}D^{\text{3D}}_{t}$, ${}^{{\text{pure}}}D^{\text{2D}}_{t}$, before-after processing}
% ${}^{{\text{pure}}}D^{\text{2D}}_{t}$ are directly forwarded to downstream modules without further processing.
% \textcolor{black}{
% To further refine the quality of mixed detections ${}^{{\text{mix}}}D_{t}$, we perform geometric alignment that leverages the spatial consistency of multi-modal data.}
This yields three categories: 3D-2D mix detections, pure 3D detections and pure 2D detections.
For pure 3D detections, we apply Score Filter (SF) and Non-Maximum Suppression (NMS) to discard unreliable results, yielding \textcolor{black}{${}^{{\text{pure}}}D^{\text{3D}}_{t}$.}
Pure 2D detections are directly forwarded to downstream modules without further processing as ${}^{{\text{pure}}}D^{\text{2D}}_{t}$.
\textcolor{black}{
To further refine the quality of mixed detections ${}^{{\text{mix}}}D_{t}$, we perform geometric alignment that leverages the spatial consistency of multi-modal data.}

\textbf{Alignment.}
% Existing methods \cite{wang2022deepfusionmot, li2023camo, kim2021eagermot}, despite seeking to leverage the spatial consistency between LiDAR and camera modalities, merely adopt a matching-only pre-processing strategy to acquire 3D-2D detection pairs.
% This amounts to an inadequate exploitation of cross-modal spatial consistency, particularly since 2D detections yield high precision.
Existing multi-modal methods \cite{wang2022deepfusionmot, li2023camo, kim2021eagermot} merely adopt a matching-only pre-processing strategy to obtain 3D-2D detection pairs.
However, they overlook the inherent planar geometric relationship between 3D bounding boxes and 2D detections.
Specifically, the projection of a 3D bounding box onto the image plane should fully enclose its corresponding 2D bounding box. 
Leveraging the higher accuracy of 2D detections, we exploit this geometric constraint to refine the 3D bounding boxes.
We propose the \textbf{Geometry-Aware Alignment Module (GAAM)} to enhance full-state spatial consistency across two modalities.
Based on the pre-established 3D-2D correspondences, this module further refines mated detection pairs ${}^{{\text{mix}}}D^{\text{3D}}_{t}$ 
by minimizing the geometric discrepancy between the projected 3D bbox and the 2D detections.

Formally, GAAM takes the mix detections ${}^{{\text{mix}}}D_{t}$ as input.
The optimization target is the 3D state \(S^\text{3D}_t = (x, y, z, w, l, h, \theta)\), which is initialized from the ${}^{{\text{mix}}}D^{\text{3D}}_{t}$.
The objective function is defined as follows:

\begin{equation}
\label{eq:process_op}
\min_{S^\text{3D}_t} \mathcal{L} = 1 - \text{IoU}(\mathcal{P}(S^\text{3D}_t), {}^{{\text{mix}}}D^{\text{2D}}_{t}),
\end{equation}

where \( \mathcal{P}(\cdot) \) 
denotes the standard 3D-2D image projection matrix, and \textcolor{black}{${}^{{\text{mix}}}D^{\text{2D}}_{t}$} represents the 2D detection in ${}^{{\text{mix}}}D_{t}$.
% Newton's method~\cite{liu1989limited} is employed to solve~\cref{eq:process_op}.
\textcolor{black}{We solve~\cref{eq:process_op} using a trust-region reflective (TRF) nonlinear least-squares method with finite-difference Jacobians~\cite{branch1999subspace}.}
This strategy yields two advantages: (1) Reduced Error.
% Minimizing the IoU discrepancy between bounding boxes provides a lower-error alternative to minimizing only the Euclidean distance between their centers. 
% Minimizing the IoU discrepancy between bounding boxes provides a bounded objective, as IoU is inherently constrained within a fixed range. 
% Moreover, IoU is a normalized metric that generalizes consistently across both 3D geometric space and the 2D image plane.
Minimizing IoU discrepancy yields a bounded objective, since IoU is constrained to a fixed range and consistently normalized across 2D image and 3D geometric spaces.
(2) Full-state Optimization.
The optimized state variables encompass the 3D geometric center, object dimensions, and heading angle, enabling a comprehensive full-state optimization scheme.
As a result, the optimization of $S^\text{3D}_t$ maintains robust numerical stability throughout the alignment process.
On frames with sync. multi-sensor data, GAAM outputs three categories: aligned 3D-2D detections ${}^{{\text{mix}}}D_{t}$, pure 3D detections ${}^{{\text{pure}}}D^{\text{3D}}_{t}$, and pure 2D detections ${}^{{\text{pure}}}D^{\text{2D}}_{t}$. 
On frames with async. single-sensor data, this module outputs single-modal detections ${}^{{\text{single}}}D^{\text{2D}}_{t}$.

% In contrast, Euclidean distance in 3D space (measured in meters) and the 2D pixel plane (measured in pixels) suffer from unit discrepancy, necessitating heuristic weighting for balance.
% Benefiting from the inherent normalization and unit consistency of IoU, our alignment avoids the instability caused by heterogeneous units between 3D space and 2D image plane.

% \begin{equation}
% \label{eq:iou}
%     IoU(B_{i}, B_{j}) = \frac{\Lambda(B_{i} \cap B_{j})}{\Lambda(B_{i} \cup B_{j})}, 
% \end{equation}

% The iteration terminates when $\lvert f(\vect{x}^{(k+1)}) - f(\vect{x}^{(k)}) \rvert < 1 \times 10^{-6}$, yielding the optimal 3D object center that replaces the original detection for correction.
% The optimization procedure utilizes the gradient $\nabla f(\vect{x}^{(k)})$ defined in Equation~\ref{eq:gradient} 
% and the Hessian matrix $\nabla^2 f(\vect{x}^{(k)})$ defined in Equation~\ref{eq:hessian}. 
% (b) Detection score preprocessing and fusion. After the detection preprocessing module, three types of detection results are generated. We calculate a fused detection score $s_{fused}$ to update the prior score $\hat{s}_{t|t-1}$:
% \begin{equation}
%     s_{\text{fused}} = \omega_{3d} \cdot s_{3d} + \omega_{2d} \cdot s_{2d}
% \end{equation}
% Where $s_{3d}$ and $s_{2d}$ are the confidence of matched 3D detection and 2D detection, respectively, the sum of $\omega_{3d}$ and $\omega_{2d}$ equals one. For single-mode detection, the score of the other mode is set to 0.

\subsection{\textcolor{black}{Frequency-Aware Cascade Matching}}
\label{sec:faca}
The matching module is designed to establish associations between trajectories and observations. 
Most previous methods~\cite{li2023camo, wang2022deepfusionmot} solely focus on sync. multi-modal data in matching, yet neglect the utilization of high-frequency trajectory-observation association. 
This oversight renders them unable to compatibly leverage multi-sensor observations across both sync. and async. timestamps. 
To address this, we propose a \textbf{Frequency-Aware Cascade Matching Module (FACM)} that \textcolor{black}{adaptively switches matching strategies based on frame synchronization status. 
FACM accommodates both data modalities and integrate various metrics for the robust construction of association between trajectories and observations.}
The processing of this module varies depending on the synchronization state of the frame.

\textbf{For Sync. Frames}, a multi-stage matching strategy is devised to effectively leverage multi-sensor observations in a compatible manner.

% ${}^{{\text{mix}}}D_{t}$
% ${}^{{\text{mix}}}D^{\text{3D}}_{t}$
% ${}^{{\text{mix}}}D^{\text{2D}}_{t}$
% pure 3D detections
% ${}^{{\text{pure}}}D^{\text{3D}}_{t}$
% pure 2D detections
% ${}^{{\text{pure}}}D^{\text{2D}}_{t}$

\textit{First Phase Mix Association (MA)}: 
Owing to the cross-modal validation and high reliability of ${}^{{\text{mix}}}D_{t}$, we prioritize matching predicted trajectories $T_{t, t-1}$ with ${}^{{\text{mix}}}D_{t}$.
Following \cite{li2024fast}, we adopt A-gIoU to measure the similarity between trajectories and detections in the global 3D coordinate system. 
The Hungarian algorithm is then employed to solve this cost matrix with category-specific thresholds $\theta_{\text{fm}}$.
The outcomes can be represented as follows: 
matched pairs $\text{MP}^1 = \{(T^\text{1m}_{t,t-1}, D^\text{1m}_{t})\}$, 
unmatched trajectories and detections $\text{UM}^1 = \{T^\text{1um}_{t,t-1}, D^\text{1um}_{t}\}$. 
% Unmatched mixed detections $D^{1um}_{mix}$ are further fed into the trajectory estimation module for trajectory initialization.

\textit{Second Phase Pure 3D Association (P3DA)}: 
Given the reliable, precise depth information of LiDAR-based detections ${}^{{\text{pure}}}D^{\text{3D}}_{t}$, we associate these detections with unmatched trajectories $T^\text{1um}_{t,t-1}$ in the second stage. 
The cost construction and solving are identical to the first stage, with the replacement of the matching thresholds $\theta_\text{sm}$.
% cost构建和solving和第一阶段一样，只是替换了matching thresold $\theta_{sm}$
% \textcolor{black}{We also employ A-gIoU to construct the cost matrix and apply the Hungarian algorithm with a matching threshold $\theta_{sm}$ for association.}
The results can be represented as follows:
matched pairs $\text{MP}^2 = \{(T^\text{2m}_{t,t-1}, D^\text{2m}_{t})\}$, 
unmatched trajectories and detections $\text{UM}^2 = \{T^\text{2um}_{t,t-1}, D^\text{2um}_{t}\}$. 
% Unmatched 3D detections $D^{2um}_{3D}$ are also fed into the trajectory estimation module for trajectory initialization.

\textit{Third Phase Pure 2D Association (P2DA)}: 
\textcolor{black}{
Despite lacking depth information, 2D detections are more robust to occlusion and can detect objects at longer ranges. 
We therefore leverage this visual modality in the third stage to reduce premature trajectory termination and guide trajectory estimation optimization.}
% We associate pure 2D detections ${}^{{\text{pure}}}D^{\text{2D}}_{t}$ with $T^{2um}_p$ in the image plane by projecting each trajectory onto the camera view using the projection matrix \( \mathcal{P}(\cdot) \) and matching them with the 2D detections.
${}^{{\text{pure}}}D^{\text{2D}}_{t}$ are associated with trajectories $T^\text{2um}_{t,t-1}$ in the image plane. 
Specifically, we first project the 3D state of each trajectory onto the camera view using the projection matrix \( \mathcal{P}(\cdot) \). 
We then match the projected boxes with the corresponding 2D detections.
2D IoU is used to construct the cost matrix, which is then solved using the Hungarian algorithm with category-specific thresholds $\theta_\text{tm}$. 
The output consists of the matched pairs $\text{MP}^3 = \{(T^\text{3m}_{t,t-1}, D^\text{3m}_{t})\}$ and the unmatched trajectories and detections $\text{UM}^3 = \{T^\text{3um}_{t,t-1}, D^\text{3um}_{t}\}$. 

% Finally, for synchronous frames, we output $\text{MP}^i, i=1, 2, 3$, $T^{3um}_p$, $D^{1um}_{t}$, $D^{2um}_{t}$, $D^{3um}_{t}$.
% third stage matched pairs $MP^3 = (T^{3m}_{p}, D^{3m}_{2D})$, unmatched trajectories and detections $UM^3 = \{T^{3um}_p, D^{3um}_{2D}\}$.
% \textcolor{black}{For all the unmatched 3D detections from $D^{1um}_{mix}$ and $D^{2um}_{3D}$, we initialize them as new trajectories, and terminate the unmatched trajectories in $T^{3um}_{p}$ whose scores fall below the delete threshold $S_d$.}

\textbf{For Async. Frames}, we extend the third association pipeline to incorporate high-frequency camera observations ${}^{{\text{single}}}D^{\text{2D}}_{t}$ across async. timestamps. 
Specifically, we associate predicted trajectories $T_{t, t-1}$ with these 2D detections. 
The resulting outputs are: 
async. matched pairs $\text{MP}^\text{a} = \{(T^\text{am}_{t,t-1}, D^\text{am}_{t})\}$, 
unmatched trajectories and detections $\text{UM}^\text{a} = \{T^\text{aum}_{t,t-1}, D^\text{aum}_{t}\}$. 

\textbf{Output.} 
For sync. frames, we output the matching pairs $\text{MP}^i$ (where $i=1,2,3$) along with the unmatched trajectories $T^\text{3um}_{t,t-1}$ and unmatched detections $D^\text{1um}_{t}$, $D^\text{2um}_{t}$, $D^\text{3um}_{t}$.
For async. frames, we output the matching pairs $\text{MP}^\text{a}$, unmatched tracklets $T^\text{aum}_{t,t-1}$ and detections $D^\text{aum}_{t}$.

\subsection{\textcolor{black}{Frequency-Aware Trajectory Estimation (Prediction)}}
\label{sec:fatep}

The trajectory estimation module first predicts both the motion and existence states of trajectories, providing essential priors for subsequent matching.
Existing methods~\cite{wang2022deepfusionmot, li2023camo, xu2024exploiting} do not effectively exploit async. observations. Incorporating such data remains challenging in current pipelines, as directly using high-frequency async. inputs often results in overconfident state estimates.
To address this limitation, we propose a \textbf{Frequency-Aware Trajectory Estimation (Prediction) Module} that operates at both sync. and async. timestamps. The module comprises motion prediction and trajectory score prediction components, both tailored for high-frequency state evolution.

\textbf{Motion Prediction.} 
We adopt the Kalman Filter (KF) as the core estimator for real-time trajectory prediction, and tailor its frame interval to accommodate the high-frequency.
% Specifically, the prediction covariance matrix $\mathbf{P}$ is optimized with distinct formulations for synchronous and asynchronous sampling, respectively.
% For synchronous sampling, it is defined as:
% \begin{equation}
% \mathbf{P}_{t,t-1} = \mathbf{F_t} \cdot \mathbf{P}_{t-1} \cdot \mathbf{F_t}^\top + \mathbf{Q},
% \end{equation}
% while for asynchronous sampling, the formulation is:
% \begin{equation}
% \mathbf{P}_{t,t-1} = \mathbf{P}_{t-1},
% \end{equation}
% where $\mathbf{F_t}$ denotes the time-adaptive state transition matrix and $\mathbf{Q}$ is the process noise covariance matrix.
% \textcolor{black}{This design differentiates the significance of synchronous and asynchronous frames, mitigating trajectory overconfidence.}

\textbf{Score Prediction.}
Following~\cite{li2024fast}, the historical trajectory score decays consistently over time:
\begin{equation}
    {s}_{t,t-1} = \sigma_\tau \cdot s_{t-1},
\end{equation}
where $s_{t-1}$ is the trajectory posterior score of frame $t-1$, $\sigma_\tau$ represents the stage-specific decay factor with $\tau \in \{\text{sync}, \text{async}\}$ (sync for sync. frame, async for async. frame), and ${s}_{t,t-1}$ is the prior score of frame $t$.
This step reflects the increased uncertainty of trajectory states over time, enabling robust subsequent trajectory updates.

% We adopt the Kalman Filter (KF) as the core estimator for real-time trajectory prediction and update. The KF is modified for high-frequency characteristics, with the prediction covariance matrix $\mathbf{P}$ optimized for synchronous sampling as:
% $$\mathbf{P}_{t|t-1} = \mathbf{F} \cdot \mathbf{P}_{t-1} \cdot \mathbf{F}^\top + \mathbf{Q},$$
% with the prediction covariance matrix $\mathbf{P}$ optimized for asynchronous sampling as:
% $$\mathbf{P}_{t|t-1} = \mathbf{P}_{t-1}$$
% where $\mathbf{F}$ is the time-adaptive state transition matrix, and $\mathbf{Q}$ is the process noise covariance.

% A key adaptive KF update strategy is designed for different timestamp states: at sync. timestamps, full KF state updates are performed using corrected 3D detections; at asynchronous timestamps, we freeze the KF update for unobserved trajectories to avoid drift. This design balances high-frequency trajectory updates with reliable state estimation, providing accurate motion states for subsequent association and lifecycle management modules.

\subsection{\textcolor{black}{Frequency-Aware Trajectory Estimation (Update)}}
\label{sec:fateu}
% For the trajectory motion and existence state update, most existing methods simply update motion states by KF and adopt a count-based existence management strategy\cite{zeng2024fusiontrack, wang2022deepfusionmot, xu2024exploiting, kim2021eagermot} \textcolor{black}{on synchronous frame.}
% % they only perform updates at timestamps with consistent Lidar and camera observations.
% In contrast, we need to handle both synchronous and asynchronous frame information \textcolor{black}{for sufficient spatio-temporal fusion.}
% A key insight is that asynchronous observations are relatively unreliable due to the lack of verification from the other modality.
% % existence 
% However, update motion states in each frame and count-based lifecycle management fail to distinguish the reliability of these two types of information. 

% Thus, to effectively differentiate between \textcolor{black}{input frames} and manage trajectories, we differentially model the observation noise to suppress unreliable motion corrections from asynchronous observations.
% % Besides, for existence states, we design a modality-informed confidence-calibrated lifecycle management strategy \textcolor{black}{to estimate the true existence probability of trajectories.}
% Besides, for existence states, we design a modality-informed and confidence-calibrated lifecycle management strategy to better estimate the true existence probability of trajectories by fusing information from both synchronous and asynchronous data.
% motion
% 我们通过对观测噪声进行差异性建模，以限制高频二维观测的不可靠矫正
The trajectory estimation module also performs robust posterior inference to produce reliable trajectory states for downstream tasks.
For motion and existence state updates, most existing methods adopt standard KF updates and count-based lifecycle management~\cite{wang2022deepfusionmot, xu2024exploiting, kim2021eagermot}, primarily designed for sync. frames. 
In contrast, Fusion-Poly integrates both sync. and async. data to achieve sufficient spatial-temporal fusion.
However, async. data is inherently less reliable due to the absence of cross-modality verification. 
Simply updating motion states at every frame or applying simple count-based management fails to account for the reliability gap between frame types.
To address this, we explicitly model observation noise in a differentiated manner and introduce a modality-aware, confidence-calibrated lifecycle management strategy. This design enables accurate trajectory state estimation by jointly leveraging sync. and async. data.

% \textcolor{black}{main existence, secondary motion.}
% Its goal is to achieve robust tracking of dynamic targets and timely elimination of invalid trajectories by dynamically adjusting the trajectory confidence score.
% Most of the existing tracking methods simply use a counting-based lifecycle management method \cite{zeng2024fusiontrack, wang2022deepfusionmot, xu2024exploiting, kim2021eagermot}. 
% They not only underutilize the rich information inherent in detection scores, but also fail to exploit valuable cues from asynchronous detection frames that are frequently available in practical multi-sensor systems.
\textbf{Motion Update.}
% noise modeling \alpha * \beta^{flag}, \beta large value, \alpha is detector-specific 
After association, the predicted trajectory motion states are updated by observations ($D_{t}^{\text{1m}}, D_{t}^{\text{2m}}, D_{t}^{\text{3m}}$, $D_{t}^{\text{am}}$). 
In the update process of KF, we heuristically model observation noise $\mathbf{R}$ from both sync. and async. frames:
\begin{equation}
    \mathbf{R} =  \gamma^{n} \cdot \mathbf{C},
\end{equation}
where $n=0$ for sync. frames and $n=1$ for async. frames, respectively.
$\gamma$ is a huge constant, and $\mathbf{C}$ is a constant diagonal matrix.

% Only the motion states of trajectories that have obtained 3D detections are updated. 
% Meanwhile, consistent with the prediction phase, the covariance of the Kalman filter is updated under synchronous frames, but not under asynchronous frames.

% (a) Trajectory score prediction. 
% The historical trajectory score decays consistently over time, using an exponential decay model:
% \begin{equation}
%     \hat{s}_{t|t-1} = \sigma \cdot s_{t-1},
% \end{equation}
% where $s_{t-1}$ is the trajectory posterior score of frame $t-1$, $\sigma$ is the decay rate, and $\hat{s}_{t|t-1}$ is the prior score of frame $t$. This step weakens the confidence of historical trajectories through the attenuation factor to symbolize the increase in uncertainty of the trajectory.

\textbf{Score Update.} 
We propose a modality-informed, confidence-calibrated lifecycle management module.
For sync. frames, the multi-modal detection scores are first fused via weighted averaging and then combined with the prior trajectory score using the Noisy-OR formulation~\cite{lemmer2004recursive} to obtain the posterior score $s_t$:

\begin{equation}
    \label{det_score_fuse}
    s_{\text{fused}} = \alpha \cdot s_{\text{3D}} + (1-\alpha) \cdot s_{\text{2D}},
\end{equation}
\begin{equation}
    s_t = 1-(1-{s}_{t,t-1})\cdot(1-s_{\text{fused}}),
    \label{eq:sync_traj_update}
\end{equation}
where $\alpha$ denotes the cross-modal fusion weight. 
%and $s_{\text{fused}}$ is the fused score.
For the async. frame, only one modality of observation confidence $s_{\text{single}}$ is available. 
The update process is formulated as:
\begin{equation}
    s_t = 1-(1-{s}_{t,t-1})\cdot(1- \beta \cdot s_{\text{single}}),
\end{equation}
where $\beta$ is an attenuation coefficient to mitigate async. uncertainty.
All update functions satisfy:
\begin{equation}
    s_t \geq \max({s}_{t,t-1}, s_{\text{o}}),
\end{equation}
where $s_{\text{o}}$ is the observation score including ${s_{\text{fused}}, s_{\text{single}}}$.
% The historical confidence and the current detection confidence are product-wise fused to avoid the overconfidence problem caused by the update.
% \textcolor{black}{After the trajectory score update, the trajectory whose online average score is lower than the deletion threshold $\theta_{\text{del}}$ is terminated.}
% After updating the trajectory score, trajectories with an online average score lower than the deletion threshold $\theta_{\text{del}}$ are terminated.
Trajectories with an online average score below $\theta_{\text{del}}$ are terminated. 
Unmatched detections $\{D_{t}^{\text{1um}}, D_{t}^{\text{2um}}\}$ are initialized as new trajectories, with their initial scores assigned following \cref{det_score_fuse}.
By differentially modeling observation noise and confidence scores, our framework effectively leverages high-frequency async. data for temporal continuity while relying on sync. frames for precision validation.

% % 为什么我们的模块可以对高低频尽量良好建模，callback 段首
% \textcolor{black}{We further perform the different update comparison experiments in the following section.}

{
\label{variance_analysis}
\textbf{Variance Analysis of Score Fusion Update.} 
% 同步的高可靠性
% We provide a theoretical justification for using the fused score $s_{\text{fused}}$ instead of the $s_{\text{3D}}$ in the trajectory quality update step. 
We theoretically justify that our differential update strategy: employing $s_{\text{fused}}$ for sync. frames and $s_{\text{single}}$ for async. frames is superior to the uniform single-modality update strategy.
$s^*$ is defined as the ground-truth existence probability of a trajectory. 
We \textcolor{black}{assume} the detection scores from each modality as noisy observations of $s$:
\begin{equation}
    s_{\text{3D}} = s^* + \epsilon_{\text{3D}}, \quad s_{\text{2D}} = s^* + \epsilon_{\text{2D}},
\end{equation}
where $\epsilon_{\text{3D}}$ and $\epsilon_{\text{2D}}$ are zero-mean noises with variances $\sigma_{\text{3D}}^2$ and $\sigma_{\text{2D}}^2$. 
Since heterogeneous sensors operate based on different physical principles and exhibit distinct failure modes, their measurement noises can be reasonably assumed to be uncorrelated.
The variance of the fused score is:
\begin{equation}
    \text{Var}(s_{\text{fused}}) = \alpha^2 \sigma_{\text{3D}}^2 + (1-\alpha)^2 \sigma_{\text{2D}}^2.
\end{equation}
This demonstrates that there exists an optimal weight $\alpha^* = \sigma_{\text{2D}}^2 / (\sigma_{\text{3D}}^2 + \sigma_{\text{2D}}^2)$ that achieves the minimum variance:
\begin{equation}
    \text{Var}(s_{\text{fused}}) = \frac{\sigma_{\text{3D}}^2 \, \sigma_{\text{2D}}^2}{\sigma_{\text{3D}}^2 + \sigma_{\text{2D}}^2},
\end{equation}
which satisfies $\text{Var}(s_{\text{fused}}) < \min(\sigma_{\text{3D}}^2, \, \sigma_{\text{2D}}^2)$. 
This indicates the fused estimate is strictly more precise than either single-modality estimate, regardless of the relative quality of the two sensors.
This variance reduction directly propagates into the trajectory score update process. 
The trajectory score is updated via~\cref{eq:sync_traj_update}: $s_t^{\text{up}} = 1 - (1 - s_t)(1 - s_{\text{input}})$, where $s_{\text{input}}$ is either $s_{\text{3D}}$ or $s_{\text{fused}}$. 
The variance of the updated score with respect to the input noise is:
\begin{equation}
    \text{Var}(s_t^{\text{up}}) = (1 - s_t)^2 \cdot \text{Var}(s_{\text{input}}).
\end{equation}
% Since $\text{Var}(s_{\text{fused}}) < \sigma_{\text{3D}}^2$, 
% replacing $s_{\text{3D}}$ with $s_{\text{fused}}$ leads to a proportional blackuction in the variance of the updated trajectory score. 
Since $\text{Var}(s_{\text{fused}}) < \text{Var}(s_{\text{3D}})$, setting $s_{\text{input}} = s_{\text{fused}}$ in \cref{eq:sync_traj_update} results in a smaller $\text{Var}(s_t^{\text{up}})$ than setting $s_{\text{input}} = s_{\text{3D}}$, i.e., a lower variance of the updated trajectory score.
% This effect accumulates over consecutive frames: a more stable input score leads to a more stable trajectory score, which further enables more reliable trajectory management decisions, such as trajectory confirmation and termination. 
% In practice, this makes the tracker less sensitive to single-frame score fluctuations caused by temporary sensor degradation, e.g., LiDAR point sparsity at long range or camera exposure artifacts, thus improving the overall robustness of the lifecycle management module.
This effect compounds over consecutive frames.

% A first-order noise propagation analysis over $T$ successive updates from an initial score $s_0$ yields:
% \begin{equation}
%     \text{Var}(s_T^{\text{fused}}) = T \cdot C_T \cdot \text{Var}(s_{\text{fused}}) < T \cdot C_T \cdot \sigma_{\text{3D}}^2 = \text{Var}(s_T^{\text{single}}),
% \end{equation}
% where $C_T = (1{-}s_0)^2(1{-}s^*)^{2(T-1)}$ is a shablack dynamic factor independent of the input modality, and the factor $T$ reflects the additive accumulation of per-frame noise. This confirms that the per-frame advantage $\text{Var}(s_{\text{fused}}) < \sigma_{\text{3D}}^2$ is preserved and compounded at the trajectory level, enabling more reliable lifecycle management decisions such as trajectory confirmation and termination.

% In practice, this makes the tracker less sensitive to single-frame score fluctuations caused by temporary sensor degradation, e.g., LiDAR point sparsity at long range or camera exposure artifacts.
}

\begin{table}[t]
\centering
\caption{Performance comparison on the nuScenes test set. CP: CenterPoint~\cite{yin2021center}; LK-F: LargeKernel-F~\cite{chen2023largekernel3d}; MV2D: Mv2DFusion~\cite{wang2025mv2dfusion}; CR : Cascade R-CNN~\cite{cai2018cascade}; BEVF: BEVFusion~\cite{liu2023bevfusion}; FC: FocalsConv~\cite{chen2022focal}.}
\label{table:nu_test}
\renewcommand{\arraystretch}{0.7}
\setlength{\tabcolsep}{1.0mm}{
\begin{tabular}{ccc|ccc}
\toprule
\multicolumn{1}{c}{\textbf{Method}} & \textbf{Detector} & \textbf{Input} & \textbf{AMOTA}$\uparrow$ & \textbf{MOTA}$\uparrow$ & \textbf{IDS}$\downarrow$ \\ \midrule
MotionTrack\cite{zhang2023motiontrack} & MotionTrack        & C + L       & 55.0      & 49.0     & 8716 \\
EagerMOT~\cite{kim2021eagermot}         & CP \& CR         & C + L       & 67.7      & 56.8        & 1156 \\
% P3D\cite{chiu2021probabilistic}         & CenterPoint~\cite{yin2021center} \& self-model         & C + L       & -      & -     & -    & -   & -   \\
Poly-MOT\cite{li2023poly}         & LK-F        & C + L       & 75.4      & 62.1     & 292 \\
CAMO-MOT\cite{li2023camo}         & BEVF \& FC        & C + L       & 75.3      & 63.5     & 324 \\
Fast-Poly\cite{li2024fast}         & LK-F        & C + L       & 75.8      & 62.8     & 326 \\
MC-Track\cite{wang2024mctrack}         & LK-F        & C + L       & 76.3      & 63.4     & \textbf{242} \\
DINO-MOT\cite{lee2024dino} & -        & C + L       & 76.3      & 65.0     & 387 \\
EMMS-MOT\cite{xu2024emmsmot}         & LK-F \& CR        & C + L       & 76.4      & -     & 271 \\
Fusion-Poly         & MV2D \& CR        & C + L       & \textbf{76.5}     & \textbf{65.1}     & 282 \\
\bottomrule
\end{tabular}}
\vspace{-1em}
\end{table}

\begin{table}[t]
\centering
\caption{Performance comparison on the nuScenes val set. CP: CenterPoint~\cite{yin2021center}; LK-F: LargeKernel-F~\cite{chen2023largekernel3d}; MV2D: Mv2DFusion~\cite{wang2025mv2dfusion}; CR: Cascade R-CNN~\cite{cai2018cascade}; BEVF: BEVFusion~\cite{liu2023bevfusion}; FC: FocalsConv~\cite{chen2022focal}; DINO: DINOv2~\cite{oquab2023dinov2}.}
\label{table:nu_val}
\renewcommand{\arraystretch}{0.7}
\setlength{\tabcolsep}{1.0mm}{
\begin{tabular}{ccc|ccc}
\toprule
\multicolumn{1}{c}{\textbf{Method}} & \textbf{Detector} & \textbf{Input} & \textbf{AMOTA}$\uparrow$ & \textbf{MOTA}$\uparrow$ & \textbf{IDS}$\downarrow$ \\ \midrule
MotionTrack\cite{zhang2023motiontrack}         & MotionTrack        & C + L       & 64.1      & 59.7     & - \\
% P3D\cite{chiu2021probabilistic}         & CP \& Self-model         & C + L       & 68.7      & -     & - \\
EagerMOT~\cite{kim2021eagermot}         & CP \& CR        & C + L       & 71.2      & -        & 899 \\
DINO-MOT\cite{lee2024dino}         & CP \& DINO      & C + L       & 73.1      & -     & 216 \\
Fast-Poly\cite{li2024fast}         & CP       & L       & 73.7      & 63.2     & 414 \\
MC-Track\cite{wang2024mctrack}         & CP        & L       & 74.0      & 64.0     & 275 \\ 
EMMS-MOT\cite{xu2024emmsmot}         & CP \& CR        & C + L       & 74.3      & -     & 247 \\
Poly-MOT\cite{li2023poly}         & CP       & L       & 73.1      & 61.9     & 232 \\
CAMO-MOT\cite{li2023camo}         & BEVF \& FC        & C + L       & 76.3      & -     & 239 \\
\textcolor{black}{Fusion-Poly}         & CP \& CR      & C + L       & \textbf{77.1}     & \textbf{67.3}     & \textbf{416} \\
\bottomrule
\end{tabular}}
\vspace{-1em}
\end{table}

%%%%%%%%%%%%%%%%%%%%%%%%%%%%%%%%%%%%%%%%%%%%%%%%%%%%%%%%%%%%%%%%%%%%%%%%%%%%%%%%

\section{EXPERIMENT}

\subsection{Dataset and Metric}

nuScenes~\cite{caesar2020nuscenes} is a large-scale autonomous driving dataset with 1,000 driving scenes. 
It equips a 360\textdegree{} sensor suite including 6 cameras (12 Hz), 5 radars (13 Hz), and 1 LiDAR (20 Hz), forming async. heterogeneous high-frequency data streams. 
Sync. data at 2 Hz is annotated to support 3D object detection, tracking, and scene understanding. 
The primary tracking metric is AMOTA~\cite{weng_ab3dmot}. 

% Evaluation Metrics

% Evaluation Metrics

% \textbf{Evaluation Metrics.}

\subsection{Implementation Details}

\textbf{Detector.} 
Fusion-Poly is implemented in Python and is learning-free. 
The multi-modal detectors used in nuScenes are listed in~\cref{table:nu_test} and~\cref{table:nu_val}.
% We project 3D boxes onto each image plane in parallel using a standard matrix to perform IoU-based matching. 

% For fair comparisons, on the nuScenes validation set, we employ CenterPoint~\cite{yin2021center} as 3D detector and Cascade R-CNN~\cite{cai2018cascade} as 2D detector.
\textbf{Multi-View Similarity.} 
We project 3D bounding boxes onto each image plane using standard projection matrices to perform IoU-based matching.
If a 3D detection is not visible in a camera, its overlap is set to zero.
For detections visible in multiple cameras, following~\cite{kim2021eagermot}, we retain only the 2D match with the largest projected area and discard the others.

\textbf{Async. Data Production.}
% Unless otherwise specified, the inputs for all experiments consist of LiDAR 3D detections at the official frequency (2 Hz) and camera 2D detections at a different frequency (4 Hz).
% We insert intermediate 2D detection between every two consecutive keyframes, yielding a 4~Hz data stream.
% Since the multiple cameras in nuScenes are not strictly time-sync. at non-keyframes, we perform temporal alignment by selecting. 
% Specifically, for each camera, we select the image whose timestamp is closest to the midpoint between keyframes and treat the resulting images as sync. observations.
% 2D detector is performed on these images to produce the intermediate 2D detections.
Unless otherwise specified, all experiments use LiDAR 3D detections and camera 2D detections at the official nuScenes keyframe frequency of 2 Hz. For camera inputs, we construct a 4 Hz 2D detection stream from the raw camera data.
Specifically, between each pair of consecutive keyframes, we select for each camera the raw image whose timestamp is closest to the temporal midpoint. These approximately aligned multi-camera images are then processed by the 2D detector to obtain intermediate detections. Together with the keyframe detections, they form the 4 Hz camera detection stream.

\textcolor{black}{\textbf{Parameter Settings.}}
We perform a linear search over the parameters to maximize AMOTA on the validation set. 
The optimized hyperparameters are then applied to the test set. 
Following~\cite{li2023poly,wang2024mctrack, li2026offline}, Fusion-Poly employs the category-specific techniques. 
A sensitivity analysis of the tuned hyperparameters is provided in Section~\ref{sec:sensitivity}. 
\textcolor{black}{For more details on hyperparameter settings, please refer to our code.}

% \textcolor{red}{high/low-freq.}
% \textcolor{red}{high-freq, multi-cam sample.}

% % For GAAM, our convergence tolerance and maximum number of iterations are $10^{-7}$ and $10^5$. 
% Please refer to our source code for detailed parameter specifications
% % Following~\cite{li2023poly, li2024fast, wang2024mctrack}, we take category-specific \textcolor{black}{strategies} for 7 categories \textcolor{black}{$(bicycle, bus, car, motorcycle, pedestrian, trailer, truck)$} in nuScenes. 
% % % 类别对应
% % For the \textcolor{black}{FACM} module, the first matching thresholds $\theta_{\text{fm}}$ are (1.55, 1.2, 1.1, 1.5, 0.9, 0.8, 1.25), the second association thresholds $\theta_{\text{sm}}$ are (0.8, 0.8, 1.5, 1.3, 1.8, 0.8, 1.2) and third association threshold $\theta_{\text{tm}}$ are (0.5, 0.9, 0.1, 0.3, 0.3, 0.1, 0.2). 
% % For the FATE module, the decay rates $\sigma_{\text{sync}}$ are set to (0.1, 0.7, 0.45, 0.7, 0.75, 0.65, 0.2), the cross-modality fusion weight $\alpha$ are (0.6, 0.8, 0.7, 0.55, 0.75, 0.65, 0.65) and delete thresholds $\theta_{\text{del}}$ are (0.05, 0.05, 0.07, 0.03, 0.15, 0.01, 0.05). 
% % For asynchronous hyperparameters, $\sigma_{\text{async}}$ are (0.7, 1.0, 0.8, 1.0, 0.7, 1.0, 0.8), the attenuation coefficients $\beta$ are (0.75, 0.75, 0.45, 0.55, 0.65, 0.05, 0.65).

\subsection{Comparative Evaluation}
As shown in \cref{table:nu_test}, Fusion-Poly achieves the SOTA performance among all TBD methods on the nuScenes test set with \textcolor{black}{76.5\%} AMOTA. 
Fusion-Poly outperforms the advanced method DINO-MOT~\cite{lee2024dino} by \textcolor{black}{0.1\%} in AMOTA, demonstrating the competitiveness of our approach.
On the nuScenes val set, Fusion-Poly also outperforms among LiDAR-Camera based methods as shown in \cref{table:nu_val}.
By employing CenterPoint~\cite{yin2021center} at 2 Hz and Cascade R-CNN~\cite{cai2018cascade} at 4 Hz, Fusion-Poly achieves 77.1\% AMOTA, outperforming CAMO-MOT~\cite{li2023camo} by 0.8\% in AMOTA. 
Furthermore, Exp. 3 in \cref{table:ablation} shows that even when tracking under sync. frames, our method still achieves 76.9\% AMOTA, demonstrating the effectiveness.

\begin{table*}[t]
\centering
\caption{Ablation experiment of Fusion-Poly on nuScenes val set. 
\textcolor{black}{Async. Data}: using 4 Hz camera-only 2D detections;
FACM: matching for sync./async. frames; 
FATE: confidence-calibrated lifecycle management;
GAAM: geometry-aware alignment.
}
\label{table:ablation}
\renewcommand{\arraystretch}{0.7}
\setlength{\tabcolsep}{2.3mm}{
\begin{tabular}{c|cccc|ccc|ccc}
\toprule
% \multicolumn{1}{c}{\textbf{Index}} & \textbf{Coordinator} & \textbf{multi-stage association} & \textbf{lifecycle} & \textbf{high freq data} & \textbf{AMOTA}$\uparrow$ & \textbf{MOTA}$\uparrow$ & \textbf{IDS}$\downarrow$ & \textbf{FN}$\downarrow$ & \textbf{FP}$\downarrow$ \\ \midrule
\multicolumn{1}{c}{\textbf{Index}} & \textbf{Async. Data} & \textbf{FACM} & \textbf{FATE} & \textbf{GAAM} & \textbf{AMOTA}$\uparrow$ & \textbf{AMOTP}$\downarrow$ &\textbf{MOTA}$\uparrow$ & \textbf{IDS}$\downarrow$ & \textbf{FP}$\downarrow$ & \textbf{FN}$\downarrow$ \\ \midrule
\textbf{Exp0}   & \textbf{\text{--}}   & \textbf{\text{--}}          & \textbf{\text{--}}          & \textbf{\text{--}}          & 75.5       & 0.557  & 65.9   & 884 & 12841 & 16804 \\
\textbf{Exp1}   & \textbf{\text{--}}   & \checkmark          & \textbf{\text{--}}          & \textbf{\text{--}}          & 76.6       & 0.543  & 66.7   & 470 & 13332 & 15935 \\
\textbf{Exp2}   & \textbf{\text{--}}   & \checkmark          & \checkmark          & \textbf{\text{--}}          & 76.8       & 0.538  & 66.7   & 443 & 13189 & 15917 \\
\textbf{Exp3}   & \textbf{\text{--}}   & \checkmark         & \checkmark           & \checkmark         & 76.9       & 0.541  & 67.0   & 446 & 13131 & 15894 \\
\textbf{Exp4}   & \checkmark   & \textbf{\text{--}}         & \textbf{\text{--}}          & \textbf{\text{--}}          & 75.4       & 0.565  & 65.6   & 886 & 12330 & 17425 \\
\textbf{Exp5}   & \checkmark   & \checkmark          & \textbf{\text{--}}          & \textbf{\text{--}}          & 76.6       & 0.557  & 67.3   & 463 & 13168 & 15678  \\
\textbf{Exp6}   & \checkmark   & \checkmark          & \checkmark          & \textbf{\text{--}}          & 77.0       & 0.546  & 67.2   & 420 & 12727 & 15967 \\
\textbf{Exp7}   & \checkmark   & \checkmark          & \checkmark          & \checkmark          & \textbf{\text{77.1}}   & \textbf{0.546}    & \textbf{\text{67.3}}  & \textbf{\text{416}}   & \textbf{\text{12679}} & \textbf{\text{15974}}  \\ 
\bottomrule
\end{tabular}}
\vspace{-1em}
\end{table*}

\begin{table}[t]
\centering
\caption{Comparison of different metric settings in GAAM on nuScenes val set. Eucl means Euclidean distance.}
\label{table:ablation_gaam}
\renewcommand{\arraystretch}{0.7}
\setlength{\tabcolsep}{2.3mm}{
\begin{tabular}{c|ccc}
\toprule
\multicolumn{1}{c}{\textbf{Metric}} & \textbf{AMOTA}$\uparrow$ & \textbf{MOTA}$\uparrow$ & \textbf{IDS}$\downarrow$ \\ \midrule
% EMMS-MOT~\cite{xu2024emmsmot} & - & - & - \\
% Eucl-low & 76.83 & 66.75 & 443 \\
% IoU-low & 76.91 & 67.03 & 446 \\
% GIoU-low & 76.80 & 66.57 & 442 \\
% \toprule
% high-freq
% Eucl & 76.99 & 67.23 & 420 \\
% GIoU & 76.99 & 67.17 & 414 \\
% IoU & \textbf{77.11} & \textbf{67.35} & \textbf{416} \\
Eucl & 77.0 & 67.2 & 420 \\
GIoU & 77.0 & 67.2 & 414 \\
IoU & \textbf{77.1} & \textbf{67.4} & \textbf{416} \\
% \toprule
% IoU-high \& Noise & 76.92 & 66.84 & 418 \\
% IoU-high \& All-Noise & 75.82 & 65.29 & 432 \\

\bottomrule
\end{tabular}}
\vspace{-1em}
\end{table}

\begin{table}[t]
\centering
\caption{Comparison of noise robustness on the nuScenes val set. 
Zero-mean Gaussian noise with varying standard deviations is injected into camera extrinsics to simulate real-world sensor noise. 
$\dagger$: We reproduced the results of EagerMOT~\cite{kim2021eagermot} using the official code. 
All methods use the same detectors: CenterPoint~\cite{yin2021center} and Cascade R-CNN~\cite{cai2018cascade}.
}
\label{table:ablation_noise}
\renewcommand{\arraystretch}{0.6}
\setlength{\tabcolsep}{1.2mm}{
\begin{tabular}{cc|ccc}
\toprule
\textbf{Method} & \textbf{Deviation} & \textbf{AMOTA}$\uparrow$ & \textbf{MOTA}$\uparrow$ & \textbf{Degradation}$\downarrow$ \\ 
\midrule
\multirow{4}{*}{EagerMOT$\dagger$}
& - & 70.3 & 60.5 & - \\
& 0.1 & 49.3 & 40.0 & -29.9\% \\
& 0.2 & 40.4 & 34.3 & -42.5\% \\
& 0.3 & 36.5 & 31.6 & -48.1\% \\
\midrule
\multirow{4}{*}{Fusion-Poly}
& - & 77.1 & 67.3 & - \\
& 0.1 & 66.5 & 57.7 & -13.8\% \\
& 0.2 & 65.0 & 56.1 & -15.7\% \\
& 0.3 & 63.8 & 55.6 & -17.3\% \\
\bottomrule
\end{tabular}}
\vspace{-1em}
\end{table}

\begin{table}[t]
\centering
\caption{Comparison of association strategy on nuScenes val set. MA, P3DA, P2DA are defined in \cref{sec:faca}.}
\label{table:asso}
\renewcommand{\arraystretch}{0.7}
\setlength{\tabcolsep}{1.5mm}{
\begin{tabular}{c|ccc|ccc}
\toprule
\multicolumn{1}{c}{\textbf{Method}} & MA & P3DA & P2DA & \textbf{AMOTA}$\uparrow$ & \textbf{MOTA}$\uparrow$ & \textbf{IDS}$\downarrow$ \\ \midrule
Exp0 & \checkmark & - & - & 75.5 & 65.6 & 809 \\
Exp1 & \checkmark & \checkmark & - & 76.5 & 66.5 & 420 \\
Exp2 & \checkmark & \checkmark & \checkmark & \textbf{77.1} & \textbf{67.3} & \textbf{416} \\
\bottomrule
\end{tabular}}
\vspace{-1em}
\end{table}

% \begin{table}[t]
% \vspace{0.3em}
% \centering
% \caption{Comparison of Lifecycle Management on nuScenes val set}
% \label{table:nu_test}
% \renewcommand{\arraystretch}{0.7}
% \setlength{\tabcolsep}{2.3mm}{
% \begin{tabular}{cc|ccc}
% \toprule
% \multicolumn{1}{c}{\textbf{Hz}} & \textbf{Hz} & \textbf{AMOTA}$\uparrow$ & \textbf{MOTA}$\uparrow$ & \textbf{IDS}$\downarrow$ \\ \midrule
% 2Hz & 2Hz & - & - & - \\
% 2Hz & 4Hz & - & - & - \\
% 10Hz & 2Hz & - & - & - \\

% \bottomrule
% \end{tabular}}
% \vspace{-1.5em}
% \end{table}

% collect multi-modal methods
\begin{figure*}
    \centering
    \includegraphics[width=1\linewidth]{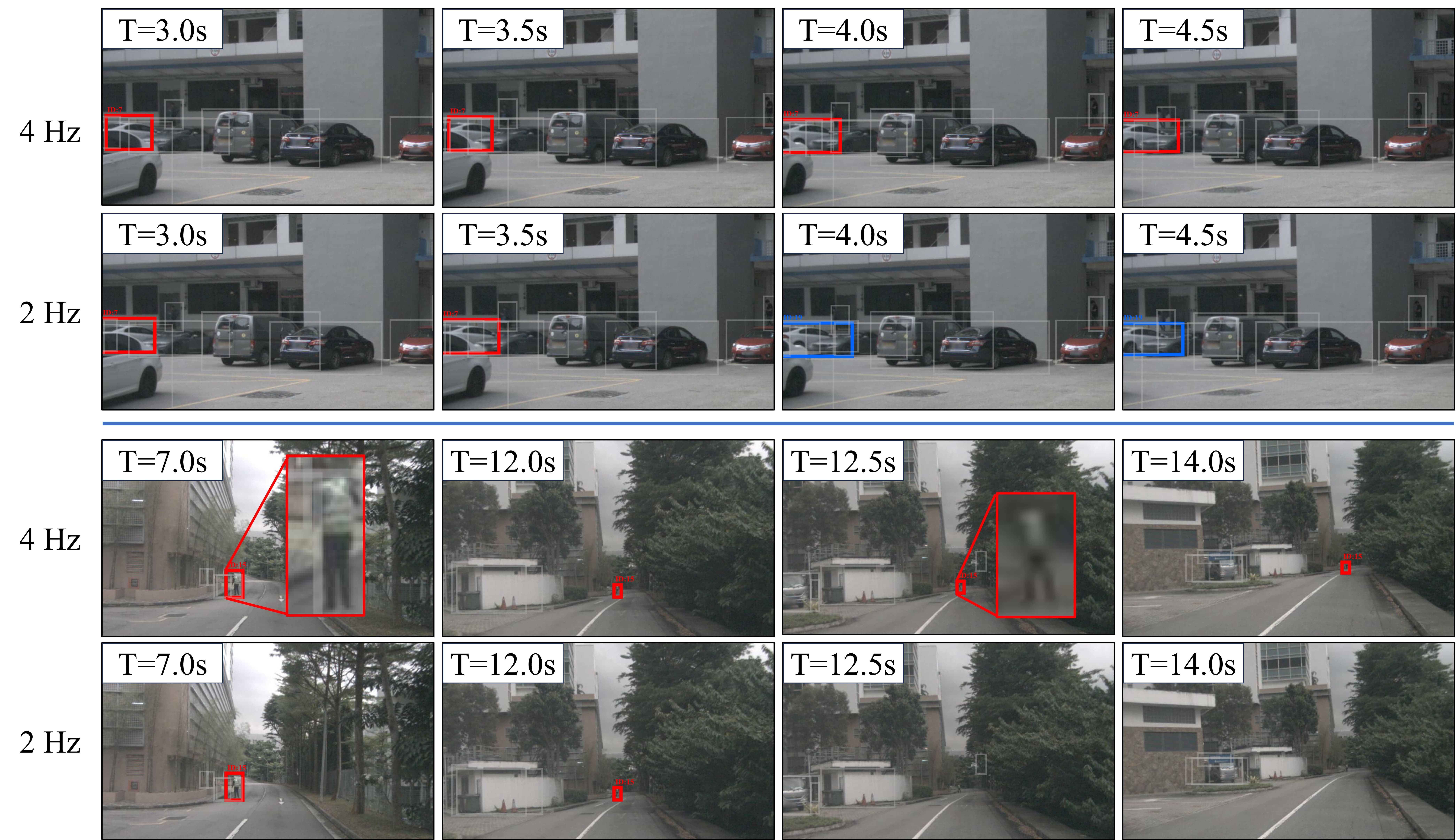}
    \caption{\textbf{Visualization results of Fusion-Poly with different inputs.} 
    In this figure, 4 Hz and 2 Hz denote the input frequency of Cascade R-CNN results, and both settings adopt 2 Hz CenterPoint results.
    }
    \label{fig:viz}
    \vspace{-1em}
\end{figure*}

\subsection{Ablation Studies} 

\textbf{Baseline Setting.} 
For \cref{table:ablation}, our baseline is configured with the following components: 
(1) Pre-processing: GAAM is not applied. 
(2) \textcolor{black}{
Matching: a single-stage approach associates 3D detections with predicted trajectories using 3D GIoU.
}
(3) Trajectory Estimation: KF and a count-based strategy are adopted, following the setting of \cite{li2023poly}.

\textcolor{black}{\textbf{Effect of Async. Data.}}
% As shown in \cref{table:ablation}, the benefit of high-frequency observations is contingent on proper frequency setting.
% Naively introducing 4~Hz 2D detections without dedicated modules (Exp.~0 vs.\ Exp.~4) degrades AMOTA by 0.1\%, indicating that async. observations damage the classic TBD framework.
% When equipped with FACM and FATE (Exp.~2 vs.\ Exp.~6), async. data yields a 0.2\% AMOTA gain, and the full configuration (Exp.~3 vs.\ Exp.~7) achieves a further 0.2\% improvement.
% Notably, adding FATE under async. input (Exp.~5 vs.\ Exp.~6) provides a 0.4\% AMOTA boost, confirming that confidence-calibrated lifecycle management is essential for distinguishing the reliability between sync and async. observations.
% These results validate our core design: async. high-frequency observations complement sync. data when frequency-aware mechanisms mitigate their inherent uncertainty.
FACM yields 1.1\% and 1.2\%
AMOTA gains under sync. and async. settings, respectively
\cref{table:ablation}. \cref{table:asso} further shows that cascade matching
improves trajectory association (+1.6\%AMOTA) and 2D
assisted existence maintenance (+0.6\%AMOTA).

\textbf{Effect of Matching Module.}
As shown in \cref{table:ablation}, multi-stage association (FACM) can effectively distinguish the detection quality of multi-modal detection results, enabling more accurate association between predicted trajectories and true detections.
When operating only on sync. frames, FACM yields a 1.1\% AMOTA improvement over single-stage association. 
Similarly, when async. data streams are introduced, it still provides a 1.2\% AMOTA gain.
\Cref{table:asso} demonstrates the effectiveness of cascade matching.
Compared with single-stage association, FACM better facilitates correct associations for predicted trajectories (+ 1.6\% AMOTA). 
Meanwhile, association with 2D detections also helps maintain the existence state of trajectories (+0.6\% AMOTA).

% FACM yields 1.1\% and 1.2\% AMOTA gains under sync. and async. settings, respectively (\cref{table:ablation}).
% \Cref{table:asso} further shows that cascade matching improves trajectory association (+1.6\% AMOTA) and 2D-assisted existence maintenance (+0.6\% AMOTA).

\textbf{Effect of Estimation Module.}
Table~\ref{table:ablation} shows that our FATE module effectively maintains trajectories under both sync. and async. data conditions. 
It improves AMOTA by 0.2\% when using only sync. frames, and by 0.4\% when incorporating async. frame data. 
\textcolor{black}{Furthermore, Exp.~0, 1, 2, 4, 5, and 6 further demonstrate that count-based lifecycle management fails to effectively distinguish the detection quality between sync. and async. data.}
Our method achieves this distinction and delivers a 0.2\% AMOTA improvement.
We also compare different lifecycle strategies in \cref{table:lifecycle}. 
Compared with other score schemes, the Noisy-OR model achieves effective fusion of 2D-3D observations, enabling flexible trajectory maintenance and termination.
% FATE improves AMOTA by 0.2\% on sync. frames and 0.4\% with async. data (\cref{table:ablation}), confirming that confidence-calibrated lifecycle management effectively distinguishes reliability between sync. and async. observations.
% \Cref{table:lifecycle} shows that the Noisy-OR model outperforms alternative score update schemes for trajectory maintenance.

\begin{table}[t]
\centering
\caption{Comparison of different score update strategies between $s_{\text{fused}}$ and $s_{t,t-1}$ on nuScenes val set. EMA: exponentially weighted moving average.}
\label{table:lifecycle}
\renewcommand{\arraystretch}{0.7}
\setlength{\tabcolsep}{2.3mm}{
\begin{tabular}{c|ccc}
\toprule
\multicolumn{1}{c}{\textbf{Method}} & \textbf{AMOTA}$\uparrow$ & \textbf{MOTA}$\uparrow$ & \textbf{IDS}$\downarrow$ \\ \midrule
\textcolor{black}{Average} & 72.9 & 61.7 & 554 \\
\textcolor{black}{EMA} & 73.8 & 62.4 & 526 \\
\textcolor{black}{Max} & 75.9 & 65.7 & 484 \\
Noisy-OR (Ours) & \textbf{77.1} & \textbf{67.3} & \textbf{416} \\

\bottomrule
\end{tabular}}
\vspace{-2em}
\end{table}

\textbf{Effect of Pre-processing Module.}
% Exp. 3 and 7 in \cref{table: ablation} demonstrate the effectiveness of IoU-based full-state optimization of GAAM, which improves AMOTA by 0.1\% under both synchronized frames and with the introduction of asynchronous frames.
% MOTA 分别也有提高了 0.3\% 和 0.1\% 
% while blackucing false positives (FP). 
% Preprocessing comparison experiments (Table IV) demonstrate its IoU-high strategy outperforms Euclidean and GIoU-based optimization, with strong noise robustness. Moreover, GAAM generates high-quality detection inputs for subsequent FACM and FATE modules, and enhancing the reliability of trajectory estimation.
% Exp. 3 and 7 in~\ref{table:ablation} validate the effectiveness of the IoU-based full-state optimization implemented in GAAM, which yields a 0.1\% improvement in AMOTA for both the synchronous frame and asynchronous frames setting. Correspondingly, MOTA is increased by 0.3\% and 0.1\% in the two settings, with a simultaneous blackuction in FP. 
Exp.~3 and 7 in \cref{table:ablation} verify the efficacy of the IoU-based full-state optimization in GAAM, which boosts AMOTA by 0.1\% for both sync. and async. frame setups, with MOTA rising by 0.3\% and 0.1\% respectively in these two settings and FP decreasing concurrently.
Further, \cref{table:ablation_gaam} demonstrates that the IoU-based strategy of GAAM outperforms Euclidean distance and GIoU-based optimization methods. % and exhibits strong robustness to noise~\cref{table:ablation_noise}. 
% Specifically, GAAM generates high-quality detection inputs for the subsequent modules, enhancing the reliability of trajectory estimation.

% design various tables.
% The GAAM realizes IoU-based full-state optimization of 3D bounding boxes to boost the spatial consistency between LiDAR and camera. 
\textbf{Sensitivity to Sensor Malfunction.}
% As shown in \cref{table:ablation_noise}, to simulate miscalibration, Gaussian noise with varying standard deviations is injected into the camera extrinsics. 
% The AMOTA of EagerMOT declines by 29.9\%--48.1\%, whereas Fusion-Poly limits degradation to 13.8\%--17.3\%. 
% This robustness stems from our unified framework with joint cross-frequency integration, which mitigates projection errors through temporal consistency and multi-modal cues. 
% Fusion-Poly consistently outperforms EagerMOT, confirming its effectiveness under sensor imperfections.
As shown in \cref{table:ablation_noise}, Gaussian noise injected into camera extrinsics degrades EagerMOT by 29.9\%--48.1\%, whereas Fusion-Poly limits degradation to 13.8\%--17.3\%, confirming robustness under sensor imperfections via joint cross-frequency integration.

\begin{figure}
    \centering
    \includegraphics[width=1\linewidth]{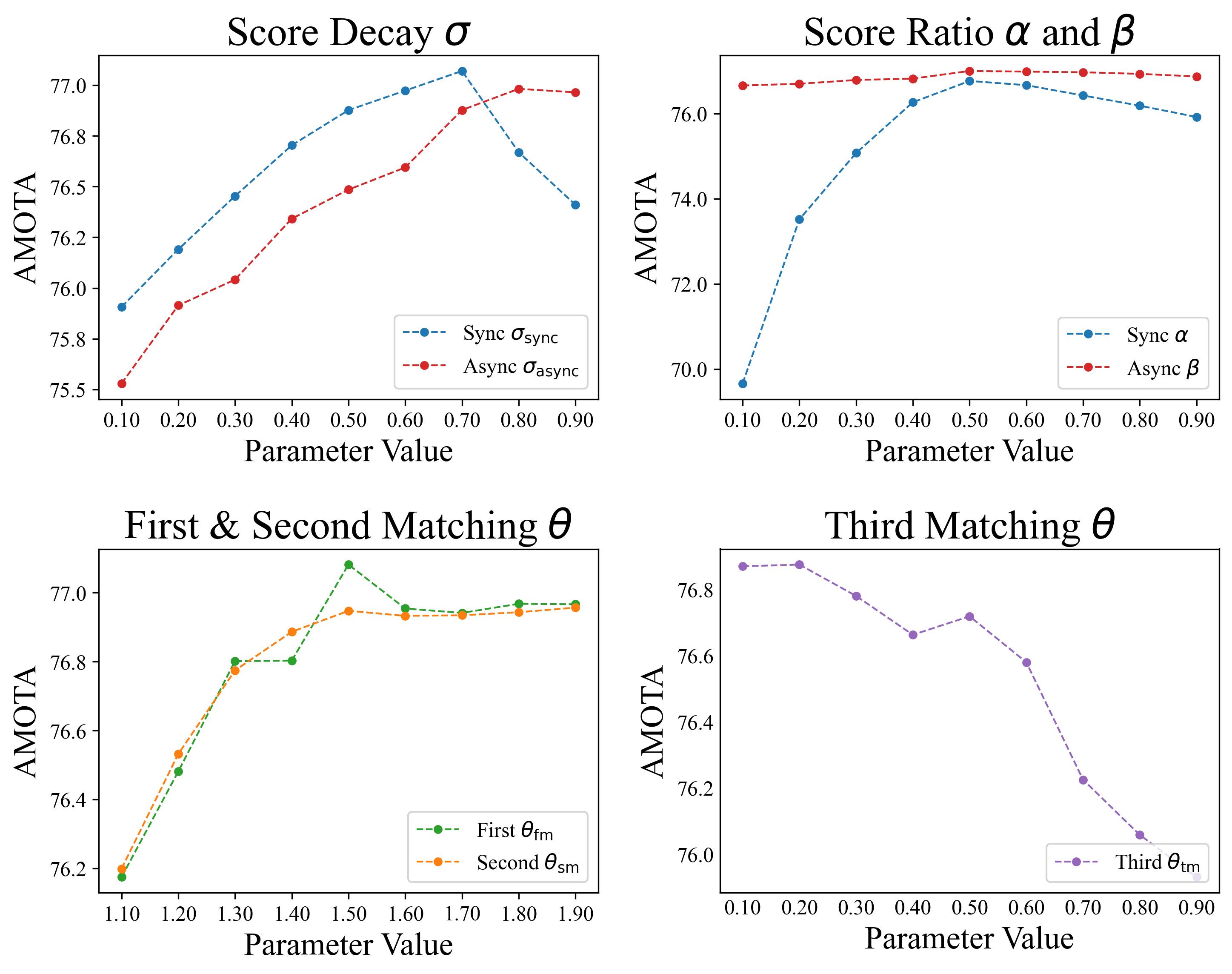}
    \caption{Performance comparison with identical key parameters across all categories.
    }
    \label{fig:hyper}
    \vspace{-2.0em}
\end{figure}

\subsection{Hyperparameter Sensitivity Analysis}
\label{sec:sensitivity}
As shown in Fig.~\ref{fig:hyper}, we evaluate the sensitivity of Fusion-Poly by performing a linear search over each hyperparameter while keeping others fixed.
For FATE, $\sigma_{\text{sync}}$ peaks around 0.7 with a concave response, as aggressive decay causes premature termination while weak decay retains stale confidence; $\sigma_{\text{async}}$ increases monotonically and saturates beyond 0.7, favoring historical confidence preservation for non-keyframe management.
The sync.\ weight $\alpha$ peaks at 0.4--0.5 with severe degradation at extremes, whereas the async.\ weight $\beta$ remains stable across all values, confirming robust async.\ score fusion.
For FACM, $\theta_{\text{fm}}$ and $\theta_{\text{sm}}$ plateau beyond 1.4 with strong insensitivity up to 1.9, while $\theta_{\text{tm}}$ peaks at small values and degrades as the threshold loosens, necessitating strict 2D association constraints.
All analyses exhibit stability within reasonable ranges, validating the robustness of Fusion-Poly.

% As shown in Fig.~\ref{fig:hyper}, we evaluate the robustness of Fusion-Poly to its core hyperparameters by performing the linear search for each parameter while keeping others fixed. 
% \textcolor{red}{The trajectory score decay factor achieves optimal AMOTA at around 0.7 and maintains stable performance within a reasonable range, whereas excessive decay leads to continuous performance degradation. 
% This indicates that the trajectory score effectively captures the uncertainty of trajectories. 
% The cross-modal score fusion weight attains peak performance in the mid-range, with significant drops under extreme single-modality bias, underscoring the importance of multi-modal score fusion across both sync. and async. timestamps.} 
% All key hyperparameters exhibit stability within reasonable ranges, validating the robustness of our framework.

% \textcolor{red}{The matching thresholds for the first two stages in FACM yield high AMOTA scores ranging from 1.4 to 1.7, demonstrating strong insensitivity to threshold variations. And the third-stage 2D matching threshold peaks at a small value, and performance degrades when the threshold becomes overly loose.}

%%%%%%%%%%%%%%%%%%%%%%%%%%%%%%%%%%%%%%%%%%%%%%%%%%%%%%%%%%%%%%%%%%%%%%%%%%%%%%%%

\section{VISUALIZATION}
% We present visualization results of Fusion-Poly with different inputs.
% As shown in \cref{fig:viz}, the top two rows show that with the introduction of asynchronous camera data, our method can effectively maintain trajectory integrity and blackuce IDS caused by jitter and ego-motion.
% The bottom two rows demonstrate that the extra detection data enables higher-frequency association and fusion, thus allowing long-range trajectories to be maintained for a longer duration.
We visualize comparative tracking results under different input frequencies in \cref{fig:viz}.
The top two rows show that in challenging parking scenarios with significant jitter, 2 Hz sync. input fails to bridge large displacement gaps, whereas 4 Hz async. data reduces prediction uncertainty and eliminates IDS.
The bottom two rows demonstrate that the increased temporal density of async. inputs provides continuous existence confirmation for long-range objects, sustaining trajectories through low-confidence periods and extending tracking duration.

% To qualitatively analyze Fusion-Poly, we visualize comparative tracking results under different input frequencies in \cref{fig:viz}.
% The top two rows highlight the capability of our method to mitigate IDS in a cluster scenario.
% \textcolor{black}{In challenging parking scenarios with significant jitter, the standard 2 Hz sync. input often fails to bridge large displacement gaps, leading to tracking discontinuity.}
% By integrating 4 Hz async. camera data, Fusion-Poly leverages high-frequency updates to reduce prediction uncertainty, enabling robust association and eliminating IDS.
% The bottom two rows demonstrate the enhanced continuity for long-range objects.
% While sparse 2 Hz observations often result in premature trajectory termination (False Negative) for weak pedestrian signals, the increased temporal density of 4 Hz async. inputs provides continuous confirmation of existence.
% This allows lifecycle management to sustain trajectories through low-confidence periods, effectively extending the tracking duration.
% These results validate that async. high-frequency data is a critical enabler for robust tracking stability.

\section{CONCLUSIONS}

% A conclusion section is not requiblack. Although a conclusion may review the main points of the paper, do not replicate the abstract as the conclusion. A conclusion might elaborate on the importance of the work or suggest applications and extensions. 
% 一句话概述Fusion-Poly + 两三句话强调其提出方法的作用及其重要性 + 一句话概括其实验效果 + 代码开源社区贡献 + 未来展望
This paper presents Fusion-Poly, a polyhedral framework based on spatial-temporal fusion for 3D MOT. 
To effectively fuse high-frequency asynchronous data and synchronous multi-modal observations, we design three key components: the frequency-aware cascade matching module enables adaptive matching for synchronous and asynchronous frames; the frequency-aware trajectory estimation module implements high-frequency differential state update with confidence-calibrated lifecycle management; the geometry-aware alignment module optimizes 3D detections to enhance cross-modal spatial consistency. 
Experiments on nuScenes demonstrate that Fusion-Poly achieves the SOTA performance, with ablation studies validating the effectiveness of each module. 
Fusion-Poly is open-source and aims to contribute to the community.
% For future work, we aim to integrate this framework with end-to-end perception pipelines or extend it to multi-sensor fusion scenarios with radars, further optimizing its performance in complex environments. 

% These components achieve robust spatio-temporal fusion of heterogeneous sensor data, markedly improving 3D MOT motion estimation accuracy and trajectory management robustness.

% \addtolength{\textheight}{-12cm}   % This command serves to balance the column lengths
%                                   % on the last page of the document manually. It shortens
%                                   % the textheight of the last page by a suitable amount.
%                                   % This command does not take effect until the next page
%                                   % so it should come on the page before the last. Make
%                                   % sure that you do not shorten the textheight too much.

%%%%%%%%%%%%%%%%%%%%%%%%%%%%%%%%%%%%%%%%%%%%%%%%%%%%%%%%%%%%%%%%%%%%%%%%%%%%%%%%
% \clearpage

% \bibliographystyle{plain}
% \bibliography{IEEEabrv, reference.bib}

% \begin{thebibliography}{99}

% \bibliographystyle{ieeetr}
% \bibliography{IEEEabrv, reference.bib}

% \end{thebibliography}

\bibliographystyle{IEEEtran}
\bibliography{reference}

\end{document}